\documentclass{article}

\usepackage[preprint]{neurips_2023}

\usepackage[utf8]{inputenc} 
\usepackage[T1]{fontenc}    
\usepackage{hyperref}       
\usepackage{url}            
\usepackage{booktabs}       
\usepackage{amsfonts}       
\usepackage{nicefrac}       
\usepackage{microtype}      
\usepackage{xcolor}         
\usepackage{graphicx} 
\usepackage{subcaption}

\title{OpusCleaner and OpusTrainer, open source toolkits for training Machine Translation and Large language models}

\author{%
  Nikolay Bogoychev \\
  School of Informatics \\
  University of Edinburgh\\
  \texttt{n.boogoych@ed.ac.uk} \\
  \And
  Jelmer van der Linde \\
  School of Informatics \\
  University of Edinburgh \\
  \texttt{jelmervdl@ed.ac.uk} \\
  \And
  Graeme Nail \\
  School of Informatics \\
  University of Edinburgh\\
  \texttt{gnail@ed.ac.uk} \\
  \And
  Barry Haddow \\
  School of Informatics \\
  University of Edinburgh\\
  \texttt{bhaddow@ed.ac.uk} \\
  \And
  Jaume Zaragoza-Bernabeu \\
  Prompsit Language Engineering \\
  SL (PLE), Spain \\
  \texttt{jzaragoza@prompsit.com} \\
  \And
  Gema Ramírez-Sánchez \\
  Prompsit Language Engineering \\
  SL (PLE), Spain \\
  \texttt{gramirez@prompsit.com} \\
  \And
  Lukas Weymann \\
  Prompsit Language Engineering \\
  SL (PLE), Spain \\
  \texttt{lweymann@prompsit.com} \\
  \And
  Tudor Nicolae Mateiu \\
  Prompsit Language Engineering \\
  SL (PLE), Spain \\
  \texttt{tudornm@prompsit.com } \\
  \And
  Jindřich Helcl\\
  Faculty of Mathematics and Physics \\
  Charles University, Czechia \\
  \texttt{helcl@ufal.mff.cuni.cz } \\
  \And
  Mikko Aulamo\\
  Department of Digital Humanities\\  Language Technology \\
  University of Helsinki, Finland \\
  \texttt{mikko.aulamo@helsinki.fi}
}

\begin{document}

\maketitle

\begin{abstract}
Developing high quality machine translation systems is a labour intensive, challenging and confusing process for newcomers to the field. We present a pair of tools \textit{OpusCleaner} and \textit{OpusTrainer} that aim to simplify the process, reduce the amount of work and lower the entry barrier for newcomers.

\textit{OpusCleaner} is a data downloading, cleaning, and proprocessing toolkit. It is designed to allow researchers to quickly download, visualise and preprocess bilingual (or monolingual) data that comes from many different sources, each of them with different quality, issues, and unique filtering/preprocessing requirements.

\textit{OpusTrainer} is a data scheduling and data augmenting tool aimed at building large scale, robust machine translation systems and large language models. It features deterministic data mixing from many different sources, on-the-fly data augmentation and more.

Using these tools, we showcase how we can use it to create high quality machine translation model robust to noisy user input; multilingual models and terminology aware models.
\end{abstract}

\section{Introduction}
\label{sec:intro}

Machine translation is ubiquitous in modern society, however training high quality machine translation systems is not trivial. A lot of the knowledge about how to build high quality systems is not well defined, comes from experience and at times may seem counter intuitive. With \textit{OpusTrainer} and \textit{OpusCleaner} we aim to explicitly address the main challenges in a user friendly manner and simplify the workload for machine translation researchers.

There are several challenges when it comes to building high quality MT systems:

\subsection{Data Sources}
Parallel data for machine translation systems comes from many different sources, that have widely varying quality. As an example, using Opus's website\footnote{\url{https://opus.nlpl.eu/}}, and filtering parallel data sources for Chinese to English, we are presented with a dozen different corpora. Here we find that  some are in traditional script, others are in simplified script, and these may or may not have been tokenized. This is before noting any language identification issues. In order to build a high quality translation system, we need to first have quality data, which necessarily means auditing each corpus manually and then deciding how to preprocess it.

\subsection{Training schedule}
High quality machine translation systems require the use of backtranslation \citep{sennrich-etal-2016-improving}, usually included in the form of pretraining. Often at the end of training models are fine tuned to in-domain data. Without a training scheduler that supports different training stages, start-and-stop training approach is necessary which presents challenge for automation and increases the burden on the researcher.

\subsection{Data Mixing}
Noisy web-crawled data is useful for translation quality, but including it too early in the training may lead to model divergence. Furthermore dirty data is orders of magnitude more available than clean manually curated parallel data. Without any upsampling, clean data might be overshadowed by dirty data, but upsampling is wasteful in terms of disk space. Finally, multilingual models require careful data mixing such that low resource language languages are not overwhelmed by high resource ones, without a training scheduler that supports data source mixing, this is achieved by upsampling low resource data and carefully mixing and shuffling it in the training data.

\subsection{Data Augmentation}
Machine translation models are training on sanitised parallel data that is usually not representative of noisy user input:
\begin{itemize}
    \item Typos are quite rare in clean data, and spellchecker is often used on web-crawled data.
    \item All caps and title-case text are often missing.
    \item Emoji are basically non existent in parallel data.
    \item Models are not trained to cope with untranslatable tokens, which should be copied between the source and the target language.
\end{itemize}

OpusTrainer and OpusCleaner are designed to resolve the above issues, and make it easy for a novice user to build high quality translation systems, by explicitly setting the expectations that training data must be carefully audited, and training data must be scheduled.

\section{OpusCleaner}
\begin{figure}[ht]
    \centering
    \includegraphics[width=0.7\textwidth]{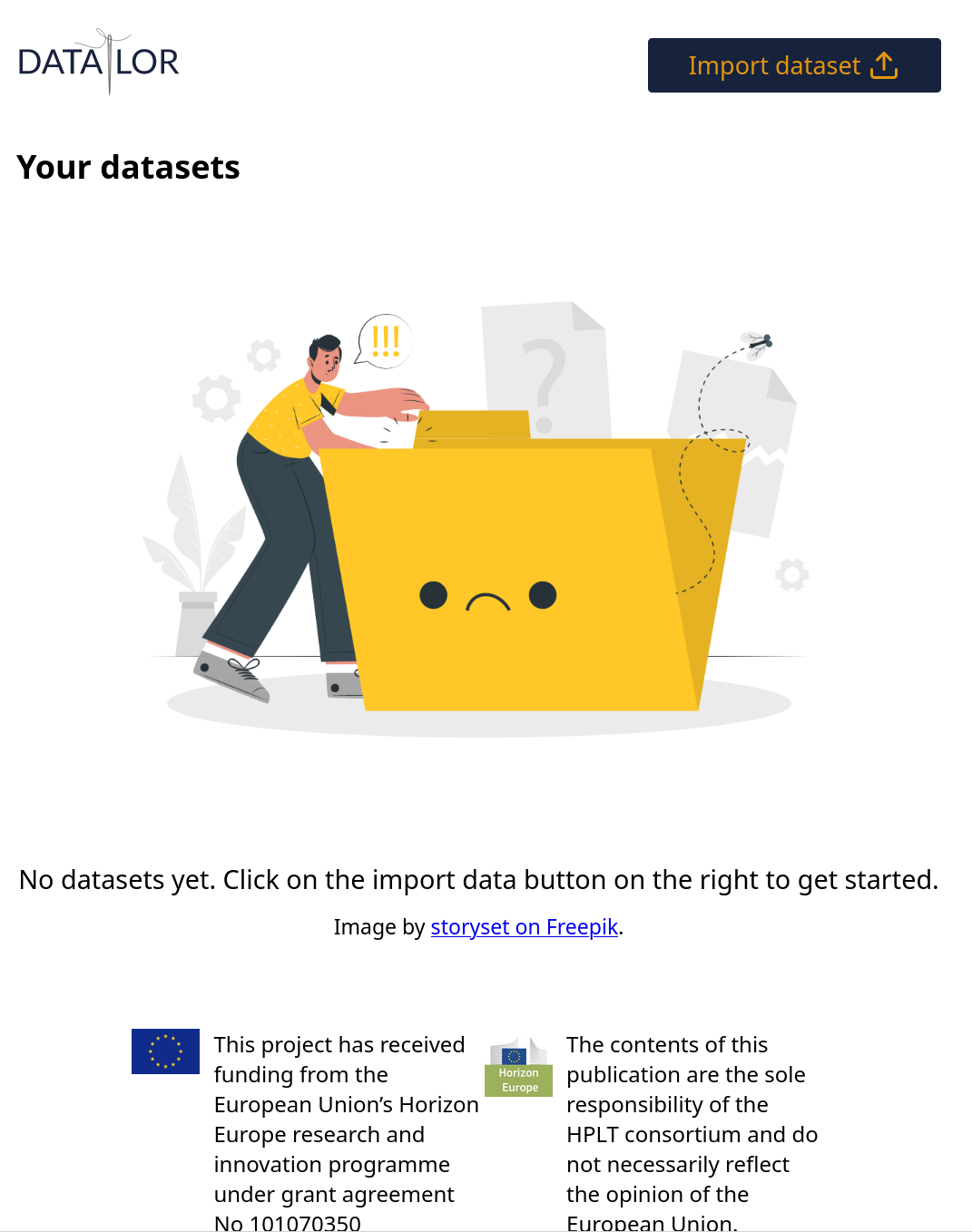}
    \caption{Initial screen of OpusCleaner}
    \label{fig:welcome_screen}
\end{figure}

In order to address the daunting task of data cleaning, we developed OpusCleaner, a single graphical frontend that does data downloading and cleaning, while being modular to allow for custom modifications depending on the language pair in question. We show screenshots of the welcome screen on Figure ~\ref{fig:welcome_screen}.

\subsection{Data Download}
\begin{figure}[ht]
    \centering
    \includegraphics[width=\textwidth]{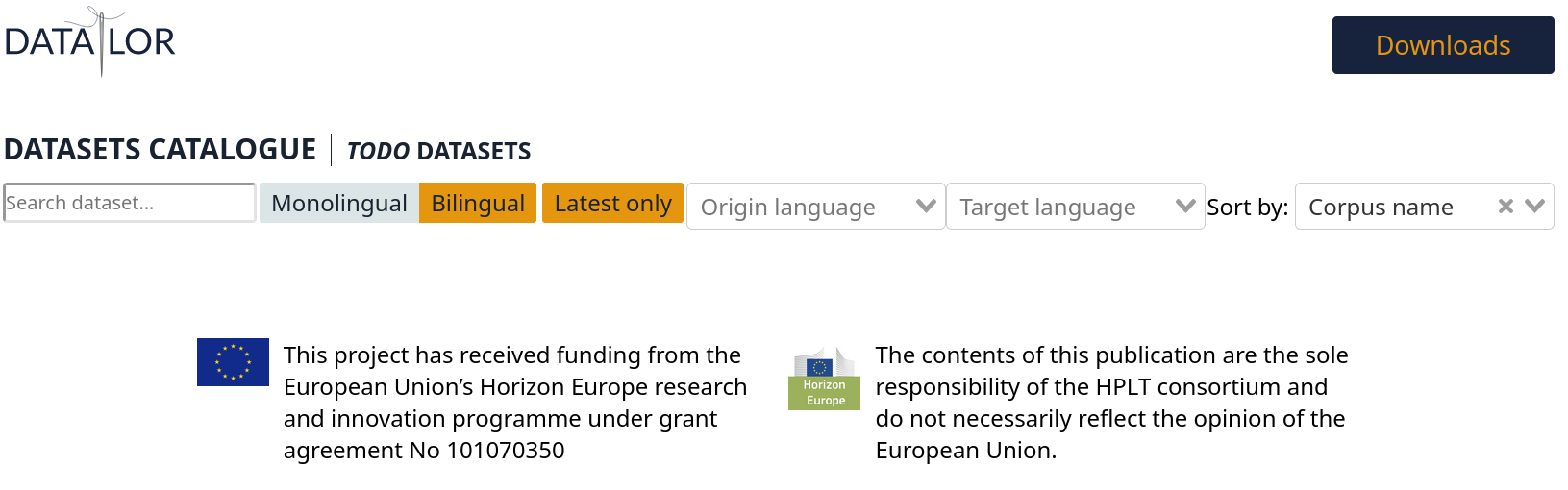}
    \caption{Search dataset pane}
    \label{fig:dataset_search}
\end{figure}

\begin{figure}[ht]
    \centering
    \includegraphics[width=\textwidth]{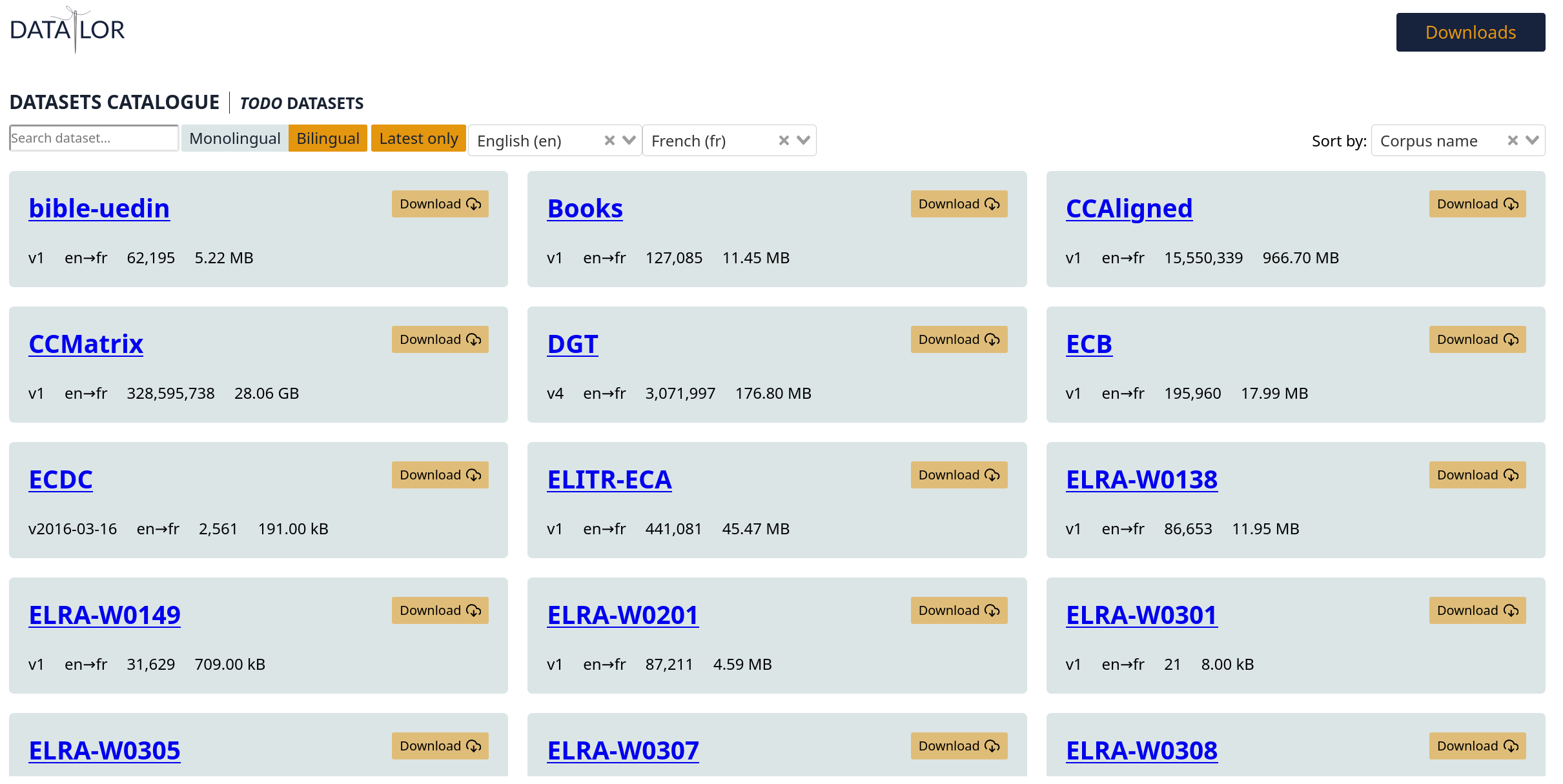}
    \caption{Search and download dataset with links to dataset and basic information.}
    \label{fig:dataset_download}
\end{figure}

OpusCleaner provides seamless integration with MTData \citep{gowda-etal-2021-many} as shown on Figures ~\ref{fig:dataset_search} and ~\ref{fig:dataset_download}. Datasets can be searched by languages and then downloaded individually, or  in bulk. Basic information about each dataset (number of lines,  version, size) are shown, as well as link to the dataset description page in Opus.

Additionally, adding one's own custom datasets is possible.

\subsection{Data Cleaning}
\begin{figure}[ht]
    \centering
    \includegraphics[width=0.7\textwidth]{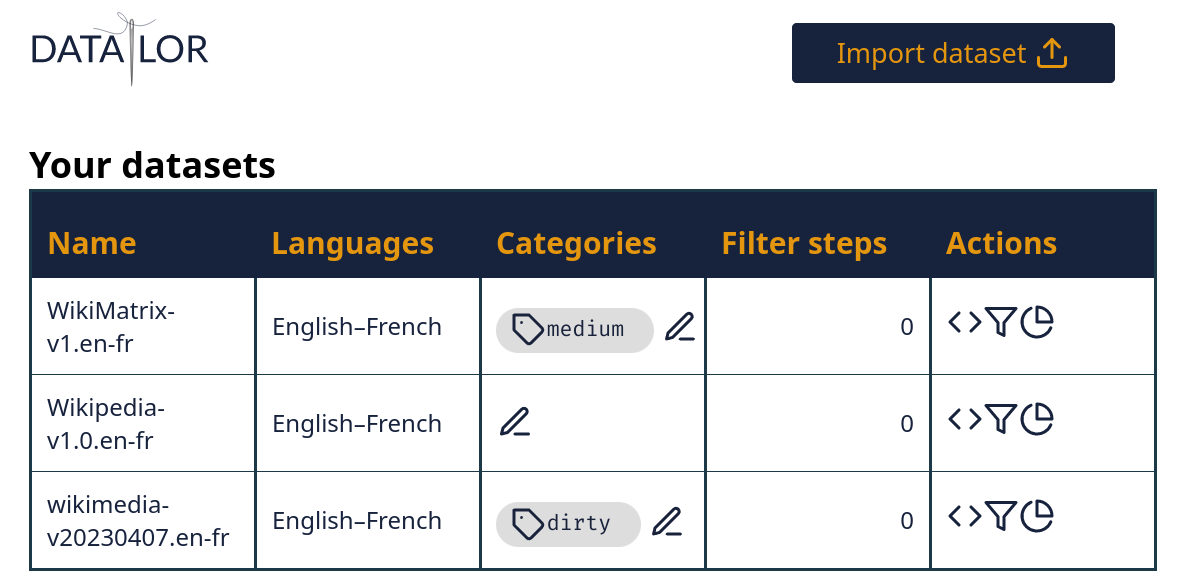}
    \caption{Initial screen of data tailoring, as well as dataset labelling.}
    \label{fig:data_tailor}
\end{figure}

Once all datasets are acquired, we can navigate to the Data Tailor screen (Figure~\ref{fig:data_tailor}) where we can label every dataset with an arbitrary label (Such as \textit{medium} or \textit{dirty}) so that we can keep track of the overall quality of each dataset.

\subsubsection{Filter and preview}
\begin{figure}[ht]
    \centering
    \includegraphics[width=\textwidth]{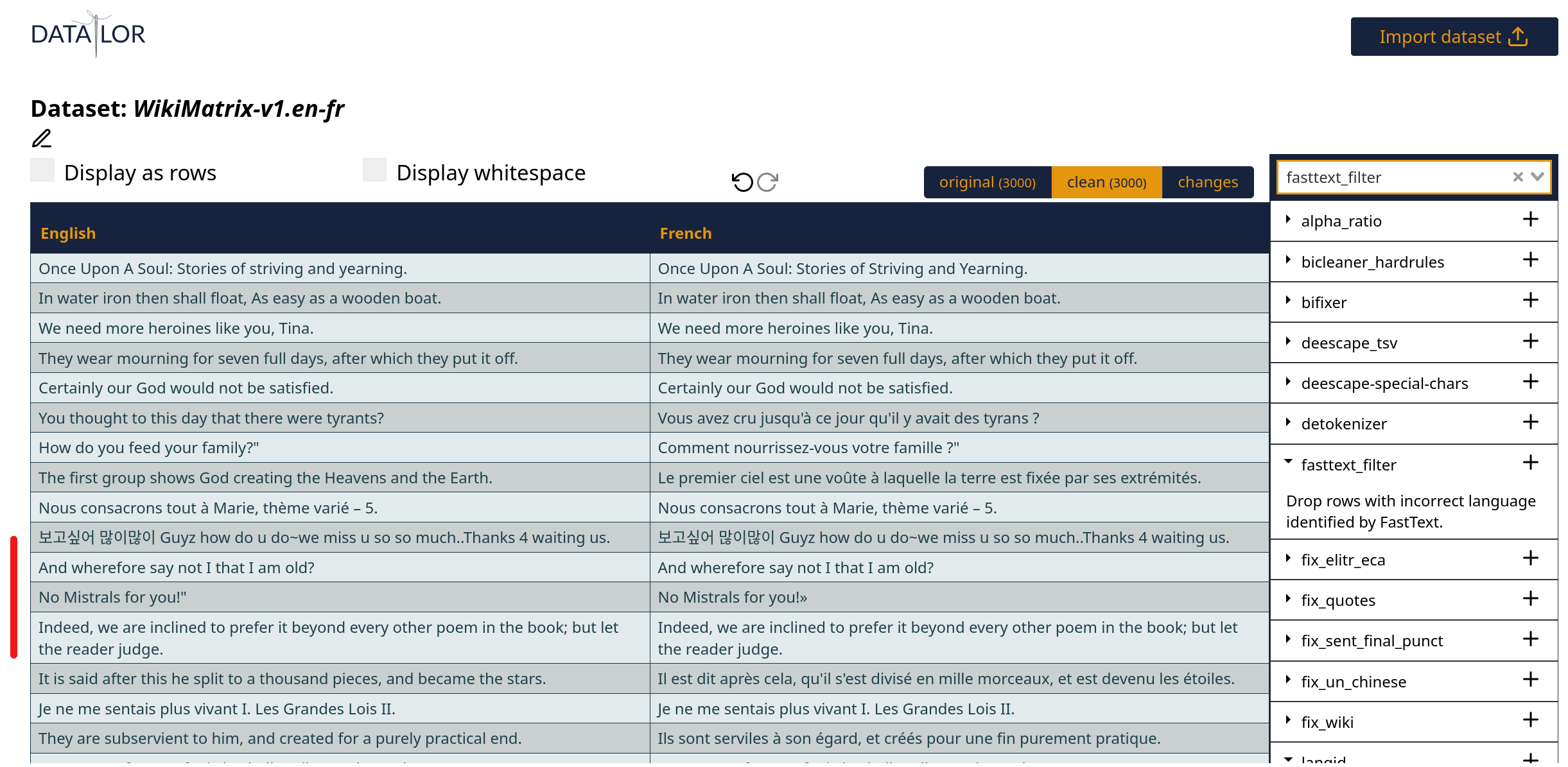}
    \caption{Initial view of dataset cleaning with some sentences obviously in the wrong language.}
    \label{fig:wrong_lang_data}
\end{figure}

\begin{figure}[ht]
    \centering
    \includegraphics[width=\textwidth]{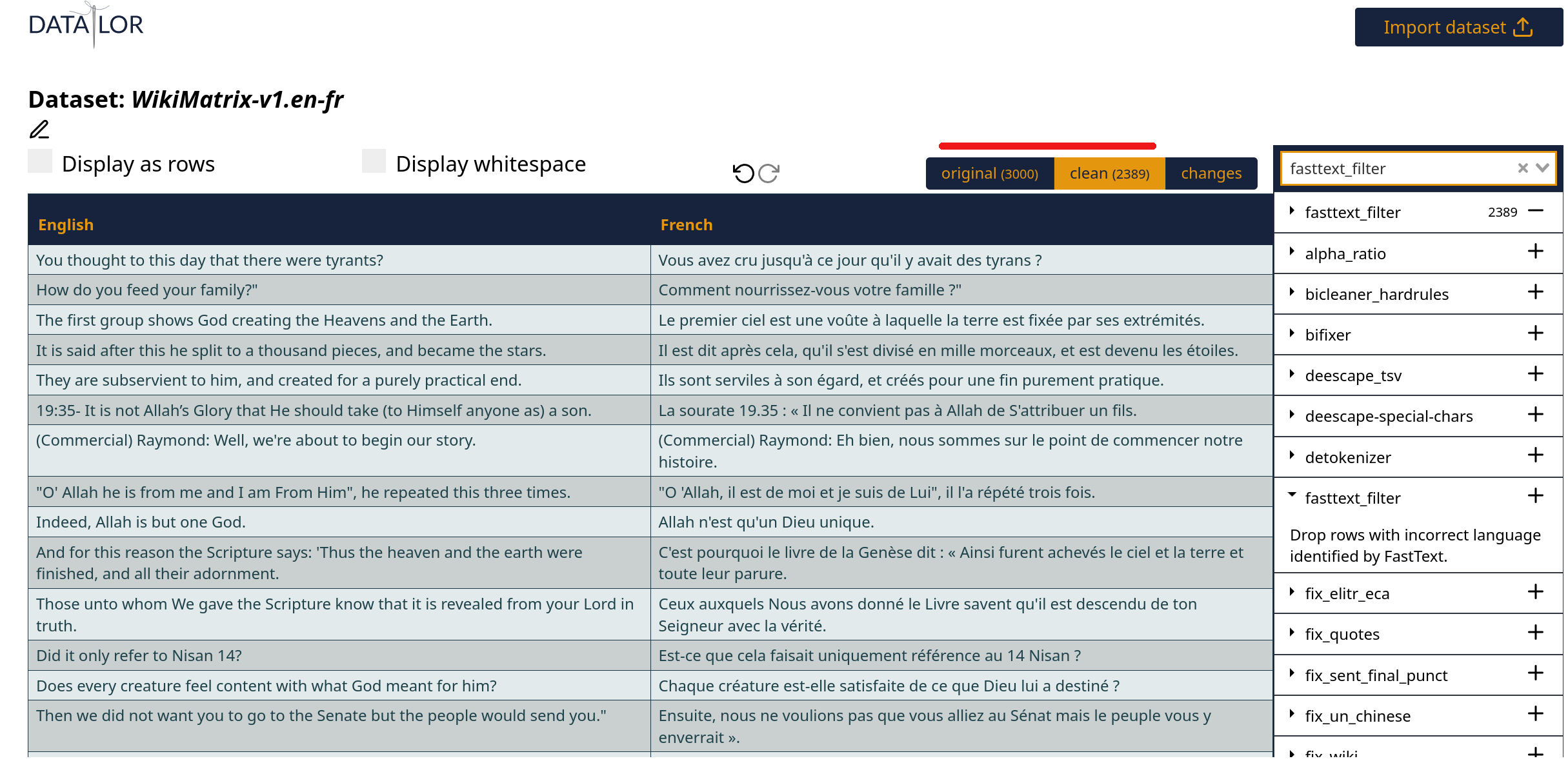}
    \caption{Fasttext langid filter removes lines in wrong language.}
    \label{fig:wrong_lang_data_fix}
\end{figure}

For each dataset, we visualise a sample of 3000 sentences that includes the first 100, the last 100 and random lines in between. From this window we can identidy the idiosyncraticies of that dataset and add the appropriate filters to fix them. For example, if we spot that some lines are in the wrong language (Figure~\ref{fig:wrong_lang_data}) we can add language identifier filter and see the result of it in the preview window (Figure~\ref{fig:wrong_lang_data_fix}).

\begin{figure}[ht]
    \centering
    \includegraphics[width=\textwidth]{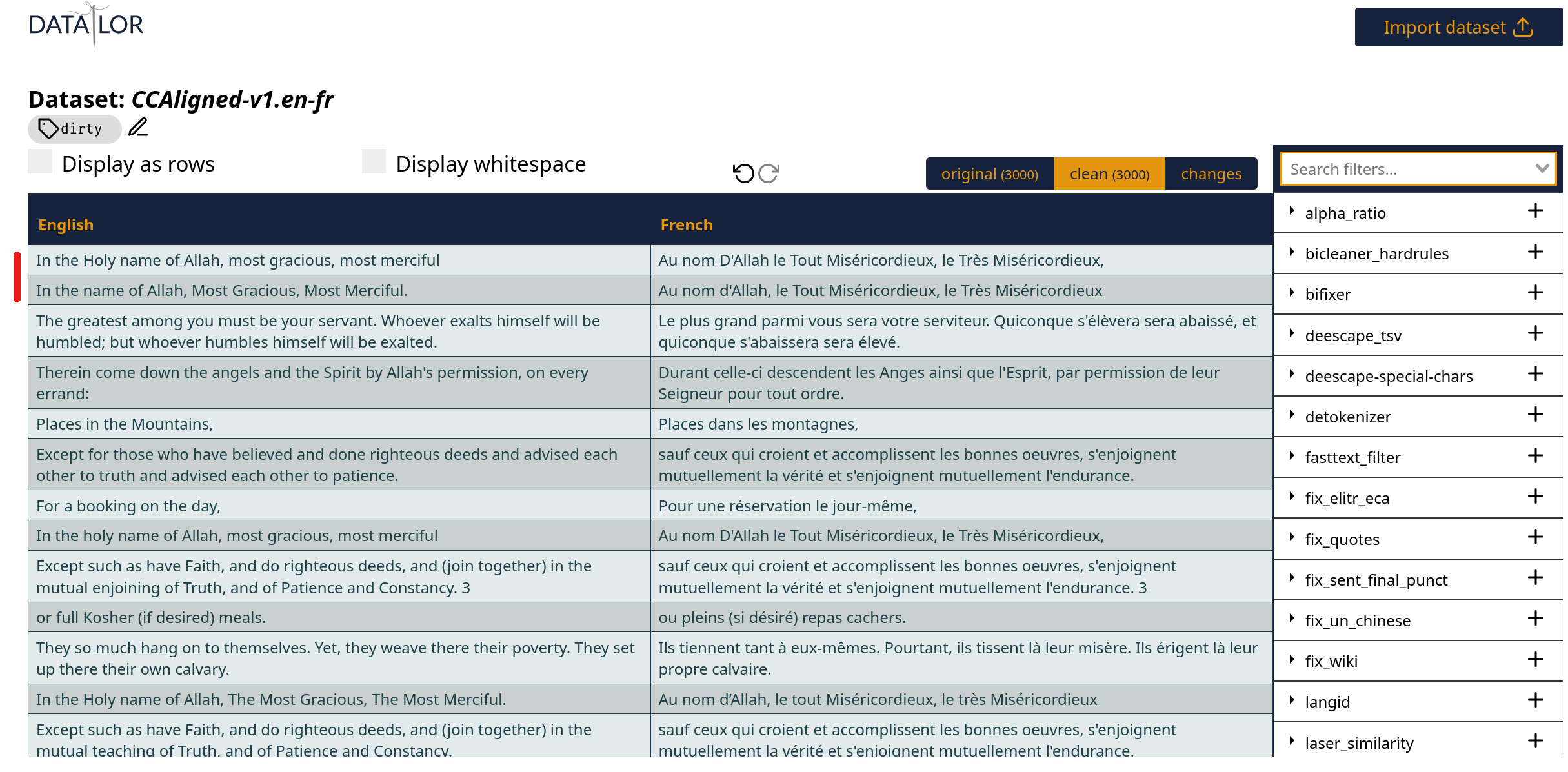}
    \caption{Mismatched punctuation on the source and the target.}
    \label{fig:mismatch_punkt}
\end{figure}

\begin{figure}[ht]
    \centering
    \includegraphics[width=\textwidth]{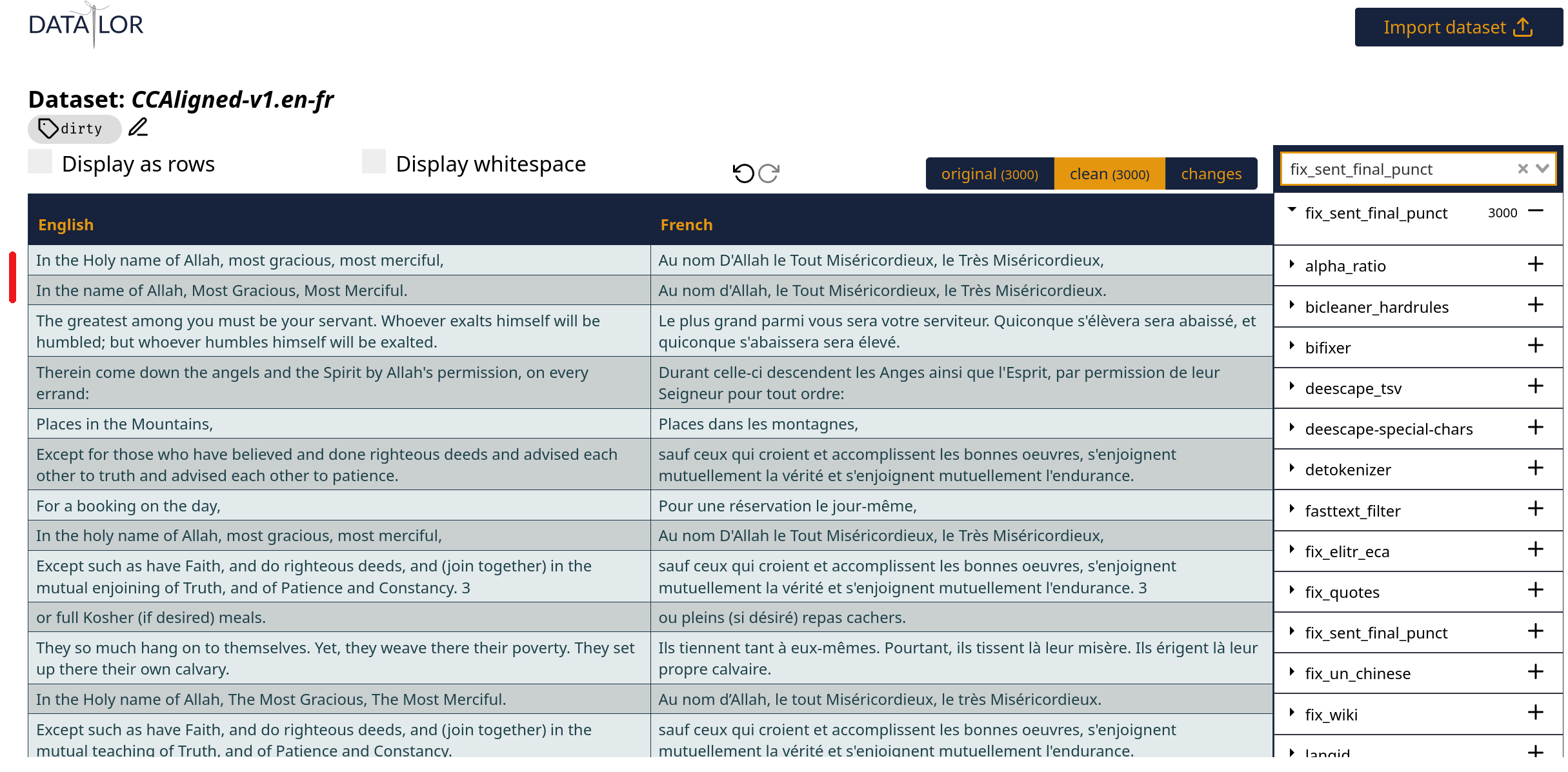}
    \caption{Fixing mismatched punctuation.}
    \label{fig:mismatch_punkt_fix}
\end{figure}

Another example is finding mismatched punctuation on the source and the target (Figure~\ref{fig:mismatch_punkt}). We can then create a simple filter that fixes the issue and apply it, see the result (Figure~\ref{fig:mismatch_punkt_fix}).

\subsubsection{Filters and pipelines}
\begin{figure}[ht]
    \centering
    \includegraphics[width=\textwidth]{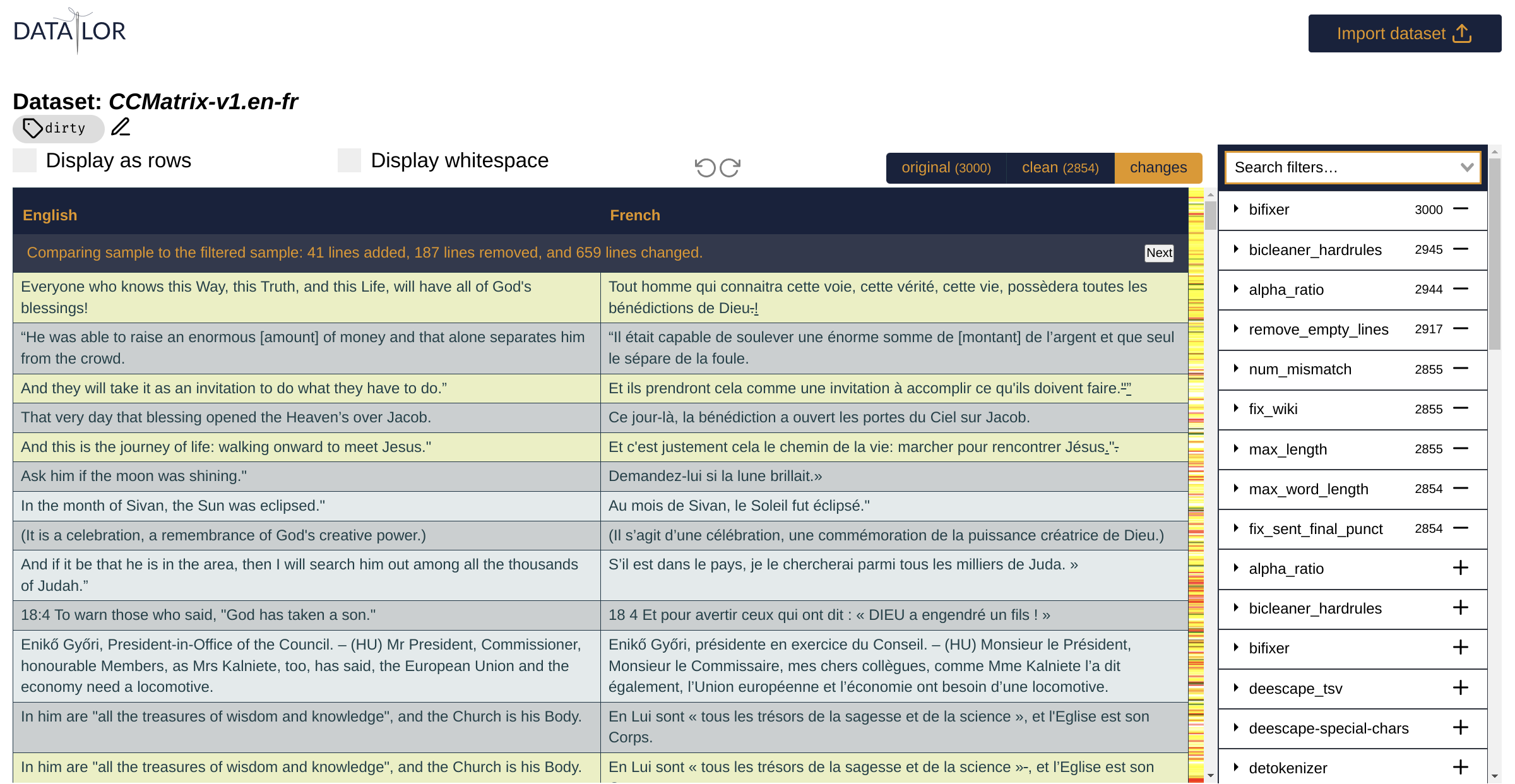}
    \caption{Adding multiple filters and visualising the difference.}
    \label{fig:filters}
\end{figure}

OpusCleaner is designed to clean data in a pipelined manner. Multiple filters are chained where every filter receives data on \textit{stdin} and outputs it on \textit{stdout}. OpusCleaner itself takes care of managing the pipeline. A typical pipeline would have a number of filters chained up as shown on Figure~\ref{fig:filters}.

We support 28 built in filters with custom user filters supported by simply providing a json configuration file that specifies path to filter executable and optionally what arguments it should have. 

\subsubsection{Processsing all data}
Once we have determined filters for every single downloaded dataset, we run a command line utility that does batch processing of all datasets, taking care of also cutting up files and parallelising processing. Once all processing is done, we provide an utility to deduplicate the data but preserving the split of datasets and then the user can proceed with training the machine translation system.

OpusCleaner\footnote{\url{https://github.com/hplt-project/OpusCleaner}} is open source, under active development and available for free for anyone to use.

\section{OpusTrainer}
As discussed in section ~\ref{sec:intro}, training high quality machine translation systems requires carefully combining parallel data from different sources and quality levels; applying on the fly modifications to it and more.

This is challenging to achieve with neural network toolkits that make use of static training data, because ideally we want to modify the data mixture and potentially augment it on the fly, without having to \textit{prepare} the data first and write it to disk which is wasteful.

\paragraph{Multilingual model training}
The problem is exacerbated when training many-to-many or English-to-many multilingual models where high resource languages would often have orders of magnitude more data than low resource languages. In order for a multilingual model to train well in this setting, it needs to see balanced data from all languages \citep{freitag-firat-2020-complete}. Doing this by concatenating and upsampling data (in order to get equal amounts of data seen for all languages), would waste multiple terabytes of disk space.

\subsection{Data Scheduling}
\begin{figure}
    \centering
    \includegraphics[width=0.7\textwidth]{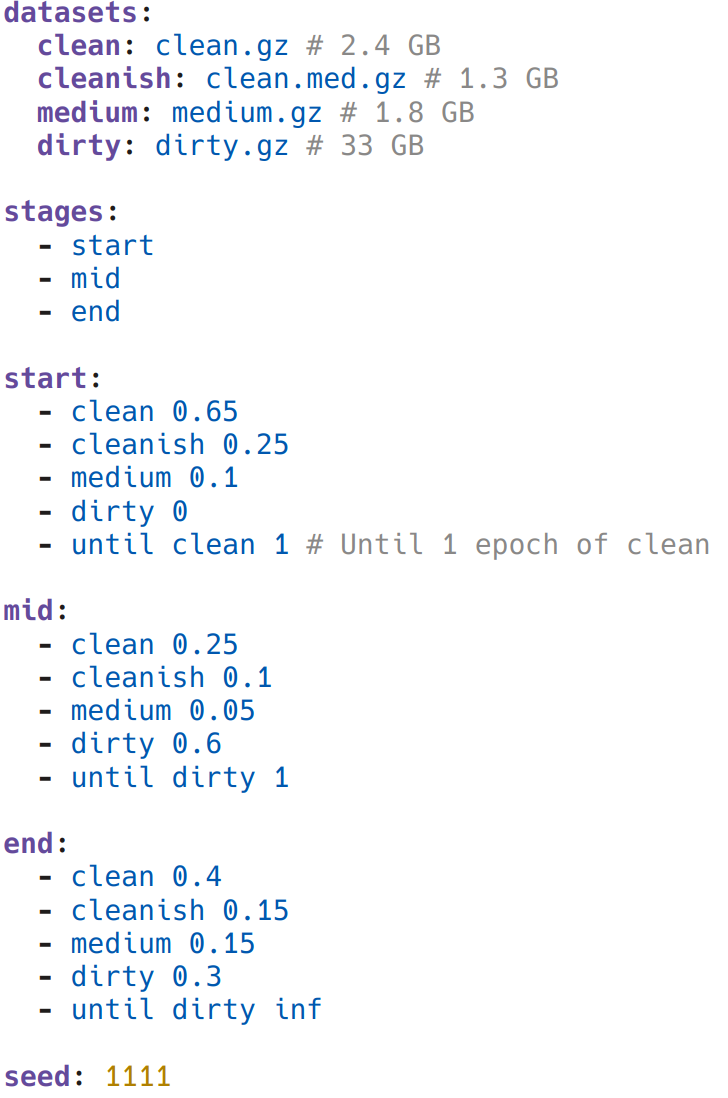}
    \caption{OpusTrainer basic configuration defining the data scheduling for training a model.}
    \label{fig:basic_opustrain}
\end{figure}

OpusTrainer solves this problem by streaming and mixing data from multiple sources. OpusTrainer uses a simple yaml configuration file where the user can declare all of their data sources and a desired mix of them for different stages of training. OpusTrainer then reads in the data from different sources and then outputs the desired mix to \textit{stdout}. OpusTrainer is meant to be used with neural network toolkits that support reading data from \textit{stdin} such as Marian \citep{mariannmt}, but it can also output the desired data mix to a file, making it usable with all toolkits. An example configuration that describes a full training run with various data mixings for different stages of training can be seen on Figure~\ref{fig:basic_opustrain}.

\subsection{Data Augmentation}
Humans are very robust to decoding noisy texts,
but this can pose a major challenge to machine translation systems due to the way we collect our training data:
\begin{itemize}
    \item Title Case and Upper Case parallel data is quite rare in training data, and is sometimes regularised during acquisition.
    \item Typos are also comparatively rare in training data, because either we use clean sources or we perform spellchecking on web crawled sources.
    \item Emojis, which human readers expect to be copied over from the source to the target, are not seen during training, because typically lines containing emojis are removed from the training data at preprocessing steps.
\end{itemize}
In order to alleviate these issues, OpusTrainer provides multiple data modifiers which can be applied on the fly, at random on the training data:

\begin{itemize}
    \item UpperCaser and TitleCaser
    \item Typo modifier, which inserts typos in words during training
    \item Merge modifier, which randomly merges several input sentences together to help the model be more robust to longer sentences.
    \item Noise modifier, that generates random sentences consisting of unicode noise, identical on both the source and the target side. This modifier teaches the model to copy unknown strings to the target side.
    \item Inline Noise modifier: A more complicated version of the above that uses word alignments in order to \textit{inject} noisy unicode characters (including Emoji) in approximately the same logical place on both the source and the target side. This modifier teaches the model that unknown sequences of \textit{<unk>} characters should be just copied on the target side.
\end{itemize}

\begin{figure}
    \centering
    \includegraphics[width=\textwidth]{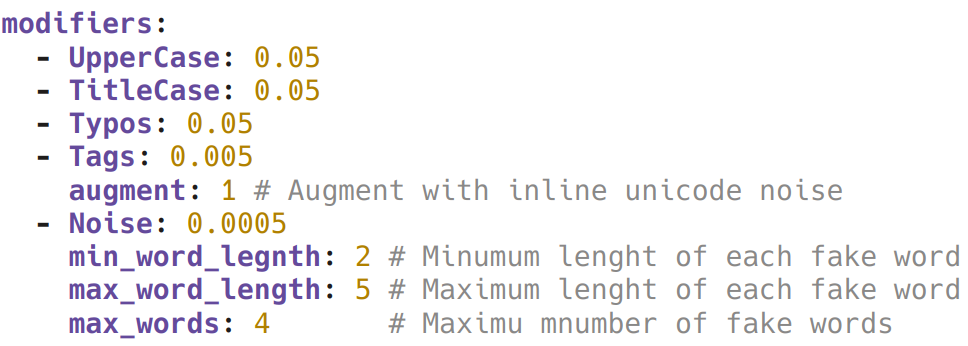}
    \caption{Different modifiers specified in YAML format to be used during training.}
    \label{fig:modifiers}
\end{figure}

All of those modifiers are applied to each sentence in the training data with a user defined probability as shown on Figure~\ref{fig:modifiers}.

\subsection{Terminology}
OpusTrainer is able to leverage word alignment information to produce terminology augmented systems, precisely as the one described in \citet{bogoychev2023terminologyaware}. This is achieved by finding bijective word alignment mappings between the source and the target sentences and at randomly injecting terminology hints in the source, precisely like the one show on ~\ref{fig:term_aug}.

\begin{figure}[htb]
        \centering
        \begin{subfigure}[t]{1.0\linewidth}
            \centering
            Where is the airport? $\leftrightarrow$ Wo ist der Flughafen? \\
Where is the airport \textit{\_\_target\_\_ Flughafen \_\_done\_\_}? $\leftrightarrow$ Wo ist der Flughafen? \\
        \end{subfigure}%
        \caption{Terminology augmentation in practise. During training it is hinted that the target word \textit{Flughafen} corresponds to \textit{Airport}, so that at inference when providing the model with terminology hints it will know how to incorporate them at the output.}
        \label{fig:term_aug}
\end{figure}

\begin{figure}
    \centering
    \includegraphics[width=\textwidth]{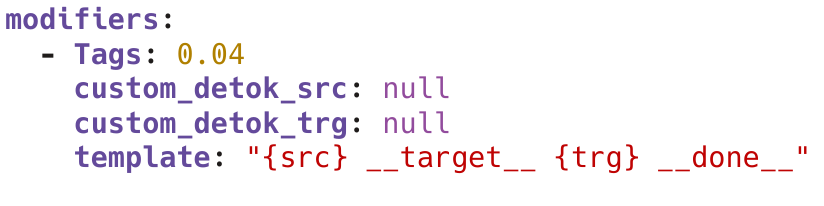}
    \caption{Tag modifier is used to add terminology hints to the source during training. Values of 3\% to 7\% seem to work well in practise.}
    \label{fig:term}
\end{figure}

These terminology hints can then be used at inference time, and the model will know how to incorporate the desired terminology hint at the target side. The relevant training options are shown on figure ~\ref{fig:term}

OpusTrainer is open source and available on GitHub,\footnote{\url{https://github.com/hplt-project/OpusTrainer}} with ample documentation and examples. OpusTrainer is designed to be used mainly with neural network toolkits that read in training input on \textit{stdin}, as it takes care of shuffling between epochs, resuming training and all other functions normally done by the data module of a neural network toolkit. It can, however, also be used to write a preprocessed training corpus on disk so toolkits that do not support reading \textit{stdin} can also make use of it.

\section{Case study: A Robust French-English system}
We highlight the use cases of data augmentation by using OpusCleaner and OpusTrainer to train a French-English machine translation system. We define robustness as the following criteria, which are all common concerns for real world web text.
\begin{itemize}
    \item Accurate translation of URLs (URLs need to be copied to the target side without any modification).
    \item Accurate copy behaviour on OOV tokens such as emoji or snippets of foreign language texts. The latter often occur in wikipedia, where foreign language terms such as named entities appear alongside their local language transliteration.
    \item No quality loss when translating Upper Case and Title case texts compared to normal cased text (All caps and tittle case often appear in tittles of newspapers).
    \item Robustness to typos (social media users).
\end{itemize}

\subsection{Test set design}

\begin{figure}
    \centering
    \includegraphics[width=\textwidth]{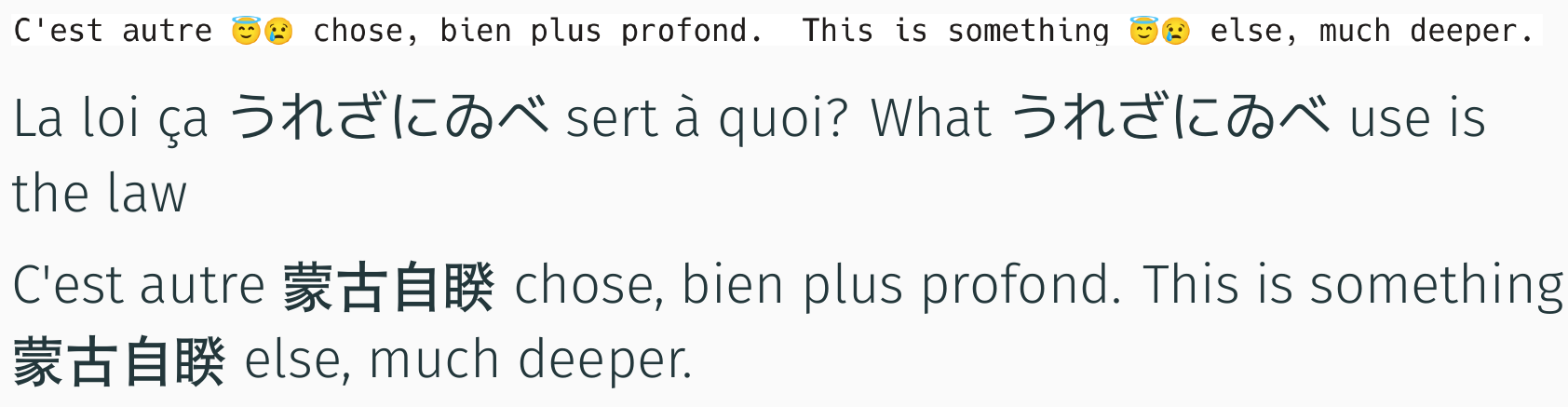}
    \caption{Example cases of noise/emoji inside the source and the corresponding target translation. We aim for our model to be able to reproduce those at decode time.}
    \label{fig:aug}
\end{figure}

As a baseline test set we use \textit{newstest15} and we make several version of it to more accurately measure robustness.
\begin{itemize}
    \item Title Case version of the test set
    \item All caps version of the test set
    \item Typo-ed version of the test set, where we insert 4 typos in each line using the python's typo library.\footnote{\url{https://pypi.org/project/typo/}}
    \item Emoji augmented test set where we insert random emoji in corresponding places on the source and the target, by using precomputed word alignments in order to place the emoji in both texts in the correct corresponding location. Example on figure ~\ref{fig:aug}.
    \item Random unicode sequence augmented test set where the random unicode sequences are inserted in the same manner as the emoji. Example on figure ~\ref{fig:aug}.
\end{itemize}

On top of that we prepare a dataset of sentences containing URLs from the paracrawl project. We take sentences containing exactly the same URLs on both the source and the target, then we remove the URLs and take the top 1500 sentences according to their bicleaner-ai \citep{zaragoza-bernabeu-etal-2022-bicleaner} score and reinsert the URLs.

For quality we report BLEU, but we also use several specific metrics. For the URL test set we measure the percentage of exact matches of URLs. For datasets with tittle case and all caps we measure as well BLEU-uncased to see how good translation quality is, regardless of the case outputted. Finally, for datasets with emoji and unicode sequences, we extract all of the OOV characters and measure ChrF \citep{popovic-2015-chrf} on them only, so that we can see how effective our system is at copying them to the target side.

\subsection{Model}
For training data we use all of the available French-English data accessible through MTData \citep{gowda-etal-2021-many} and we clean it using OpusCleaner.

We split the data into four categories based on its providence and subjective perceived quality through manual inspection:
\begin{itemize}
    \item Canonically clean datasets such as Europarl, Un are designated as clean (22M parallel sentences).
    \item Slightly less clean data (9M), designated as clean\textit{ish}.
    \item Not clean data, but not generated from crawled sources (16M), designated as medium.
    \item Web crawled data is designated as dirty (363M)
\end{itemize}

We use Marian \citep{mariannmt} to train transformer-big \citet{vaswani_transformer} models on the training data with varying degree of data augmentation. We train 7 different models with various additional \textit{perks}, some related to data augmentation, some not in order to show how we progressively achieve a more robust model.

\begin{enumerate}
    \item Pure model
    \item + Sentencepiece sampling \citep{kudo-richardson-2018-sentencepiece}. Sentencepiece sampling makes splits of words non-deterministic, potentially making unseen words handling more robust.
    \item + UpperCase and LowerCase
    \item + typos
    \item + Unicode Vocabulary Fallback. Sentencepiece models can't split OOV tokens such as Chinese characters into subwords, but if we consider that every character is represented by unicode bytes, we can split unseen characters such as emoji and hanzi
    \item + noisy sentences
    \item + inline noise
\end{enumerate}
\phantom{Hex} \\

\subsection{Results}
We present our results on table ~\ref{tab:results}. We train 7 different systems with different degrees of augmentation. We can see that progressively, as we add more modifiers to the training set up, the model becomes more robust to various sources of noisy user input. System 3 onwards have capture TitleCase and UpperCase with relatively small performance loss compared to plain sentences. System 5 that uses UTF fallback for OOV tokens starts capturing emoji and other OOV tokens. Systems 6 and 7 enhance the training data with lots of noisy examples and that leads to really good copy rate of OOV tokens to the target side, as shown in the two ChrF columns.

\begin{table*}[ht]
\small
\begin{tabular}{@{}rrrrrrrrrrrrl@{}}
\toprule
\multicolumn{1}{l}{}  & \multicolumn{10}{c}{newstest15 BLEU}                                                                                                                                                                                                                                                                                                                                                                                                                                             & \multicolumn{1}{l}{}                                                   & \multicolumn{1}{l}{}                                                             \\ \midrule
\multicolumn{1}{l}{}  & \multicolumn{1}{l}{plain} & \multicolumn{1}{c}{\begin{tabular}[c]{@{}c@{}}TC\\ uncased\end{tabular}} & \multicolumn{1}{l}{TC} & \multicolumn{1}{l}{\begin{tabular}[c]{@{}l@{}}CAPS\\ uncased\end{tabular}} & \multicolumn{1}{l}{CAPS} & \multicolumn{1}{l}{typo} & \multicolumn{1}{l}{noise} & \multicolumn{1}{c}{\begin{tabular}[c]{@{}c@{}}noise\textsuperscript{1}\\ chrf\end{tabular}} & \multicolumn{1}{l}{emoji} & \multicolumn{1}{c}{\begin{tabular}[c]{@{}c@{}}emoji\textsuperscript{1}\\ chrf\end{tabular}} & \multicolumn{1}{c}{\begin{tabular}[c]{@{}c@{}}url\\ BLEU\end{tabular}} & \multicolumn{1}{c}{\begin{tabular}[c]{@{}c@{}}URL only\\ precision\end{tabular}} \\
baseline (1)          & 40                        & 34.2                                                                     & 8.6                    & 21.5                                                                       & 20.5                     & 29.6                     & 34.3                      & 0                                                                         & 35.8                      & 0.1                                                                       & 62.7                                                                   & 90\%                                                                             \\
+ spm sample (2)      & 39                        & 36.9                                                                     & 9.1                    & 29.2                                                                       & 21.2                     & 30.5                     & 33.4                      & 0.1                                                                       & 34.7                      & 0.2                                                                       & 61.4                                                                   & 87\%                                                                             \\
+ UC/LC noise (3)     & 38.4                      & 37.3                                                                     & 36.3                   & 34.5                                                                       & 34.5                     & 29.7                     & 32.9                      & 0.1                                                                       & 34.3                      & 0.2                                                                       & 60.9                                                                   & 87\%                                                                             \\
+ typos (4)           & 38.9                      & 38                                                                       & 36.8                   & 35.1                                                                       & 35.1                     & 36.7                     & 33.5                      & 0.1                                                                       & 34.2                      & 5.2                                                                       & 61.2                                                                   & 86\%                                                                             \\
+  UTF-8 fallback (5) & 38.5                      & 38                                                                       & 36.8                   & 34.7                                                                       & 34.7                     & 36.8                     & 35.2                      & 55.1                                                                      & 37                        & 64.9                                                                      & 61                                                                     & 85\%                                                                             \\
+ noise (6)           & 39.6                      & 39.1                                                                     & 37.9                   & 35.9                                                                       & 35.9                     & 37.6                     & 38.9                      & 87                                                                        & 38.7                      & 72.3                                                                      & 61.3                                                                   & 86\%                                                                             \\
+ inline noise (7)    & 39.2                      & 38.3                                                                     & 37.2                   & 35.3                                                                       & 35.3                     & 37.5                     & 41.5                      & 92                                                                        & 39.9                      & 80.7                                                                      & 61.2                                                                   & 86\%                                                                             \\ \bottomrule
\end{tabular}
\caption{Results table \\
\textsuperscript{1} ChrF score was calculated on the noise/emoji only, meaning we only measure how well our model copies just OOV tokens without considering translation quality.}
\label{tab:results}
\end{table*}

\subsubsection{Caveats}
There are some caveats that come with our test results. The more modifiers are used, the more \textit{difficult} the training data seems to be to model, and therefore it takes more iterations through the training data to achieve convergence. Therefore all models presented have seen different amounts of training data. We will control for this setting in future work.

Furthermore we see slight degradation in terms of translation quality when we add modifications to the training data on the plain test set. This suggests that the gains we have are not entirely for free. Finally, we observe slight deterioration on URLs. We measure only exact matches on URLs because an almost correct URL is not useful. This regression bodes for further investigation.

\section{Conclusion}
We present a feature complete data preprocessing and data scheduling toolkit for training machine translation systems (but also just as useful for Large Language Models). Our tools are designed with novice and experts in mind so that they lower the entry barrier to the field of machine translation, while still allowing for state of the art results. Our data augmentation utilities are crucial for producing robust machine translation systems, as well as terminology systems \citep{bogoychev2023terminologyaware}. Our toolkit was developed concurrently and independently to Sotastream \citep{post2023sotastream} and provides similar functionality.

\bibliographystyle{plainnat}
\bibliography{custom}

\begin{thebibliography}{10}
\providecommand{\natexlab}[1]{#1}
\providecommand{\url}[1]{\texttt{#1}}
\expandafter\ifx\csname urlstyle\endcsname\relax
  \providecommand{\doi}[1]{doi: #1}\else
  \providecommand{\doi}{doi: \begingroup \urlstyle{rm}\Url}\fi

\bibitem[Bogoychev and Chen(2023)]{bogoychev2023terminologyaware}
Nikolay Bogoychev and Pinzhen Chen.
\newblock Terminology-aware translation with constrained decoding and large language model prompting, 2023.

\bibitem[Freitag and Firat(2020)]{freitag-firat-2020-complete}
Markus Freitag and Orhan Firat.
\newblock Complete multilingual neural machine translation.
\newblock In Lo{\"\i}c Barrault, Ond{\v{r}}ej Bojar, Fethi Bougares, Rajen Chatterjee, Marta~R. Costa-juss{\`a}, Christian Federmann, Mark Fishel, Alexander Fraser, Yvette Graham, Paco Guzman, Barry Haddow, Matthias Huck, Antonio~Jimeno Yepes, Philipp Koehn, Andr{\'e} Martins, Makoto Morishita, Christof Monz, Masaaki Nagata, Toshiaki Nakazawa, and Matteo Negri, editors, \emph{Proceedings of the Fifth Conference on Machine Translation}, pages 550--560, Online, November 2020. Association for Computational Linguistics.
\newblock URL \url{https://aclanthology.org/2020.wmt-1.66}.

\bibitem[Gowda et~al.(2021)Gowda, Zhang, Mattmann, and May]{gowda-etal-2021-many}
Thamme Gowda, Zhao Zhang, Chris Mattmann, and Jonathan May.
\newblock Many-to-{E}nglish machine translation tools, data, and pretrained models.
\newblock In \emph{Proceedings of the 59th Annual Meeting of the Association for Computational Linguistics and the 11th International Joint Conference on Natural Language Processing: System Demonstrations}, pages 306--316, Online, August 2021. Association for Computational Linguistics.
\newblock \doi{10.18653/v1/2021.acl-demo.37}.
\newblock URL \url{https://aclanthology.org/2021.acl-demo.37}.

\bibitem[Junczys-Dowmunt et~al.(2018)Junczys-Dowmunt, Grundkiewicz, Dwojak, Hoang, Heafield, Neckermann, Seide, Germann, Fikri~Aji, Bogoychev, Martins, and Birch]{mariannmt}
Marcin Junczys-Dowmunt, Roman Grundkiewicz, Tomasz Dwojak, Hieu Hoang, Kenneth Heafield, Tom Neckermann, Frank Seide, Ulrich Germann, Alham Fikri~Aji, Nikolay Bogoychev, Andr\'{e} F.~T. Martins, and Alexandra Birch.
\newblock Marian: Fast neural machine translation in {C++}.
\newblock In \emph{Proceedings of ACL 2018, System Demonstrations}, pages 116--121, Melbourne, Australia, July 2018. Association for Computational Linguistics.
\newblock URL \url{http://www.aclweb.org/anthology/P18-4020}.

\bibitem[Kudo and Richardson(2018)]{kudo-richardson-2018-sentencepiece}
Taku Kudo and John Richardson.
\newblock {S}entence{P}iece: A simple and language independent subword tokenizer and detokenizer for neural text processing.
\newblock In \emph{Proceedings of the 2018 Conference on Empirical Methods in Natural Language Processing: System Demonstrations}, pages 66--71, Brussels, Belgium, November 2018. Association for Computational Linguistics.
\newblock \doi{10.18653/v1/D18-2012}.
\newblock URL \url{https://aclanthology.org/D18-2012}.

\bibitem[Popovi{\'c}(2015)]{popovic-2015-chrf}
Maja Popovi{\'c}.
\newblock chr{F}: character n-gram {F}-score for automatic {MT} evaluation.
\newblock In Ond{\v{r}}ej Bojar, Rajan Chatterjee, Christian Federmann, Barry Haddow, Chris Hokamp, Matthias Huck, Varvara Logacheva, and Pavel Pecina, editors, \emph{Proceedings of the Tenth Workshop on Statistical Machine Translation}, pages 392--395, Lisbon, Portugal, September 2015. Association for Computational Linguistics.
\newblock \doi{10.18653/v1/W15-3049}.
\newblock URL \url{https://aclanthology.org/W15-3049}.

\bibitem[Post et~al.(2023)Post, Gowda, Grundkiewicz, Khayrallah, Jain, and Junczys-Dowmunt]{post2023sotastream}
Matt Post, Thamme Gowda, Roman Grundkiewicz, Huda Khayrallah, Rohit Jain, and Marcin Junczys-Dowmunt.
\newblock Sotastream: A streaming approach to machine translation training, 2023.

\bibitem[Sennrich et~al.(2016)Sennrich, Haddow, and Birch]{sennrich-etal-2016-improving}
Rico Sennrich, Barry Haddow, and Alexandra Birch.
\newblock Improving neural machine translation models with monolingual data.
\newblock In Katrin Erk and Noah~A. Smith, editors, \emph{Proceedings of the 54th Annual Meeting of the Association for Computational Linguistics (Volume 1: Long Papers)}, pages 86--96, Berlin, Germany, August 2016. Association for Computational Linguistics.
\newblock \doi{10.18653/v1/P16-1009}.
\newblock URL \url{https://aclanthology.org/P16-1009}.

\bibitem[Vaswani et~al.(2017)Vaswani, Shazeer, Parmar, Uszkoreit, Jones, Gomez, Kaiser, and Polosukhin]{vaswani_transformer}
Ashish Vaswani, Noam Shazeer, Niki Parmar, Jakob Uszkoreit, Llion Jones, Aidan~N. Gomez, \L{}ukasz Kaiser, and Illia Polosukhin.
\newblock Attention is all you need.
\newblock In \emph{Proceedings of the 31st International Conference on Neural Information Processing Systems}, NIPS'17, page 6000–6010, Red Hook, NY, USA, 2017. Curran Associates Inc.
\newblock ISBN 9781510860964.

\bibitem["Zaragoza-Bernabeu et~al.("2022")"Zaragoza-Bernabeu, Ram{\'\i}rez-S{\'a}nchez, Ba{\~n}{\'o}n, and Ortiz~Rojas]{zaragoza-bernabeu-etal-2022-bicleaner}
Jaume "Zaragoza-Bernabeu, Gema Ram{\'\i}rez-S{\'a}nchez, Marta Ba{\~n}{\'o}n, and Sergio" Ortiz~Rojas.
\newblock "bicleaner {AI}: Bicleaner goes neural".
\newblock In \emph{"Proceedings of the Thirteenth Language Resources and Evaluation Conference"}, pages "824--831", "Marseille, France", June "2022". "European Language Resources Association".
\newblock URL \url{"https://aclanthology.org/2022.lrec-1.87"}.

\end{thebibliography}


\end{document}